\documentclass[a4paper]{article}
\usepackage{INTERSPEECH2016,amssymb,amsmath,epsfig}
\ninept

\def\reg{{\rm\ooalign{\hfil
     \raise.07ex\hbox{\scriptsize R}\hfil\crcr\mathhexbox20D}}}
\title{Segmental Recurrent Neural Networks for End-to-end Speech Recognition}

\makeatletter
\def\name#1{\gdef\@name{#1\\}}
\makeatother \name{{\em Liang Lu$^{1*}$, Lingpeng Kong$^{2*}$, Chris Dyer$^2$, Noah A. Smith$^3$, and Steve Renals$^{1}$}
\thanks{$^*$ Equal contribution.  Lu and Renals are funded by the UK EPSRC Programme Grant EP/I031022/1, Natural Speech Technology (NST). The NST research data collection may be accessed at http://datashare.is.ed.ac.uk/handle/10283/786.}}

\address{$^1$Centre for Speech Technology Research, The University of Edinburgh, Edinburgh, UK \\
$^2$School of Computer Science, Carnegie Mellon University, Pittsburgh, USA\\
$^3$Computer Science \& Engineering, The University of Washington, Seattle, USA \\
{\small \tt \{liang.lu, s.renals\}@ed.ac.uk, \{lingpenk, cdyer\}@cs.cmu.edu, nasmith@cs.washington.edu}}

\begin{document}
%\ninept
%
\maketitle

\begin{abstract}
We study the segmental recurrent neural network for end-to-end acoustic modelling. This model connects the segmental conditional random field (CRF) with a recurrent neural network (RNN) used for feature extraction. Compared to most previous CRF-based acoustic models, it does not rely on an external system to provide features or segmentation boundaries. Instead, this model marginalises out all the possible segmentations, and features are extracted from the RNN trained together with the segmental CRF. Essentially, this model is self-contained and can be trained end-to-end. In this paper, we discuss practical training and decoding issues as well as the method to speed up the training in the context of speech recognition. We performed experiments on the TIMIT dataset. We achieved 17.3\% phone error rate (PER) from the first-pass decoding --- the best reported result using CRFs, despite the fact that we only used a zeroth-order CRF and without using any language model. 

\end{abstract}

\section{Introduction}
\label{sec:intro}

Speech recognition is a typical sequence to sequence transduction problem, i.e., given a sequence of acoustic observations, the speech recognition engine decodes the corresponding sequence of words or phonemes. A key component in a speech recognition system is the acoustic model, which computes the conditional probability of the output sequence given the input sequence. However, directly computing this conditional probability is challenging due to many factors including the variable lengths of the input and output sequences. The hidden Markov model (HMM) converts this sequence-level classification task into a frame-level classification problem, where each acoustic frame is classified into one of the hidden states, and each output sequence corresponds to a sequence of hidden states. To make it computationally tractable, HMMs usually rely on the conditional independence assumption and the first-order Markov rule --- the well-known weaknesses of HMMs~\cite{gillick2011don}. Furthermore, the HMM-based pipeline is composed of a few relatively independent modules, which makes the joint optimisation nontrivial. 

There has been a consistent research effort to seek architectures to replace HMMs and overcome their limitation for acoustic modelling, e.g., ~\cite{ostendorf1996,smith2001speech, gunawardana2005hidden, hifny2009speech}; however these approaches have not yet improved speech recognition accuracy over HMMs. In the past few years, several neural network based approaches have been proposed and demonstrated promising results. In particular, the connectionist temporal classification (CTC) ~\cite{graves2014towards, Hannun2014Deep, sak2015fast, miao2015eesen} approach defines the loss function directly to maximise the conditional probability of the output sequence given the input sequence, and it usually uses a recurrent neural network to extract features. However, CTC simplifies the sequence-level error function by a product of the frame-level error functions (i.e., independence assumption), which means it essentially still does frame-level classification. It also requires the lengths of the input and output sequence to be the same, which is inappropriate for speech recognition. CTC deals with this problem by replicating the output labels so that a consecutive frames may correspond to the same output label or a {\it blank} token. 
%Another problem is that CTC also replies on the frame-independence assumption, and the problem is usually mitigated by using recurrent neural networks (RNNs) for feature extraction. 

Attention-based RNNs have been demonstrated to be a powerful alternative sequence-to-sequence transducer, e.g., in machine translation ~\cite{bahdanau2014neural}, and speech recognition ~\cite{chorowski2015attention, lu2015study, chan2015listen}. A key difference of this model from HMMs and CTCs is that the attention-based approach does not apply the conditional independence assumption to the input sequence. Instead, it maps the variable-length input sequence into a fixed-size vector representation at each decoding step by an attention-based scheme (see ~\cite{bahdanau2014neural} for further explanation). It then generates the output sequence using an RNN conditioned on the vector representation from the source sequence. The attentive scheme suits the machine translation task well, because there may be no clear alignment between the source and target sequence for many language pairs. However, this approach does not naturally apply to the speech recognition task, as each output token only corresponds to a small size window of acoustic spectrum. 

In this paper, we study segmental RNNs~\cite{kong2015segmental} for acoustic modelling. This model is similar to CTC and attention-based RNN in the sense that an RNN encoder is also used for feature extraction, but it differs in the sense that the sequence-level conditional probability is defined using an segmental (semi-Markov) CRF~\cite{sarawagi2004semi}, which is an extension on the standard CRF~\cite{lafferty2001conditional}. There have been numerous works on CRFs and their variants for speech recognition, e.g, ~\cite{gunawardana2005hidden, hifny2009speech, zweig2011speech} (see \cite{fosler2013conditional} for an overview). In particular, feed-forward neural networks have been used with segmental CRFs for speech recognition~\cite{abdel2013deep, he2015segmental}. However, segmental RNNs are different in that they are end-to-end models --- they do not depend on external systems to provide segmentation boundaries and features, instead, they are trained by marginalising out all possible segmentations, while the features are derived from the encoder RNNs, which are trained jointly with the segmental CRFs. Our experiments were performed on the TIMIT dataset, and we achieved 17.3\% PER from first-pass decoding with zeroth-order CRF and without using any language model --- the best reported result using CRFs. 
 
%Perviously, but it usually requires ad-hoc rules to extract segmental-level feature representations This work is also closely related to previous works on combing the segmental CRF with feedforward neural networks~\cite{abdel2013deep, he2015segmental}, where some ad-hoc rules were used to extract segment-level feature representations.

% In this paper, we show that it is straightforward to use an RNN to extract segmental level features for CRFs.  

\section{Segmental Recurrent Neural Networks}

%In this section, we first review segmental CRFs, and then introduce using an encoder RNN approach for feature extraction, which yields the segmental RNN. We then discuss the training and decoding algorithms for this model. 

\subsection{Segmental Conditional Random Fields}

Given a sequence of acoustic frames $\mathbf{X} = \{\mathbf{x}_1, \cdots, \mathbf{x}_T\}$ and its corresponding sequence of output labels $\mathbf{y} = \{y_1, \cdots, y_J\}$, where $T\ge J$, segmental (or semi-Markov) conditional random field defines the sequence-level conditional probability with the auxiliary segment labels $\mathbf{E} = \{\mathbf{e}_1, \cdots, \mathbf{e}_J\}$ as
\begin{align}
\label{eq:crf}
P(\mathbf{y}, \mathbf{E} \mid \mathbf{X}) = \frac{1}{Z(\mathbf{X})} \prod_{j=1}^J \exp f \left( y_j, \mathbf{e}_j, \mathbf{X} \right),
\end{align}
where $\mathbf{e}_j = \langle s_{j}, n_{j} \rangle$ is a tuple of the beginning ($s_{j}$) and the end ($n_{j}$) time tag for the segment of $y_j$, and $n_j > s_j $ while $n_j, s_j \in [1, T]$; $y_j \in \mathcal{Y}$ and $\mathcal{Y}$ denotes the vocabulary set; $Z(\mathbf{X})$ is the normaliser that that sums over all the possible $(\mathbf{y, E})$ pairs, i.e.,
\begin{align}
Z(\mathbf{X}) = \sum_{\mathbf{y,E}} \prod_{j=1}^J \exp f \left( y_j, \mathbf{e}_j, \mathbf{X} \right).
\end{align}
Here, we only consider the zeroth-order CRF, while the extension to higher order models is straightforward. Similar to other CRF-based models, the function $f(\cdot)$ is defined as
\begin{align}
\label{eq:phi}
f \left( y_j, \mathbf{e}_j, \mathbf{X} \right) = \mathbf{w}^\top \Phi (y_j, \mathbf{e}_j, \mathbf{X}),
\end{align}
where $\Phi(\cdot)$ denotes the feature function, and $\mathbf{w}$ is the weight vector. Previous works on CRF-based acoustic models mainly use heuristically handcrafted feature function $\Phi(\cdot)$. They also usually rely on an external system to provide the segment labels. In this paper, we define $\Phi(\cdot)$ using neural networks, and the segmentation $\mathbf{E}$ is marginalised out during training, which makes our model self-contained. 

\subsection{Feature Representations}

\begin{figure}[t]
\small
\centerline{\includegraphics[width=0.45\textwidth]{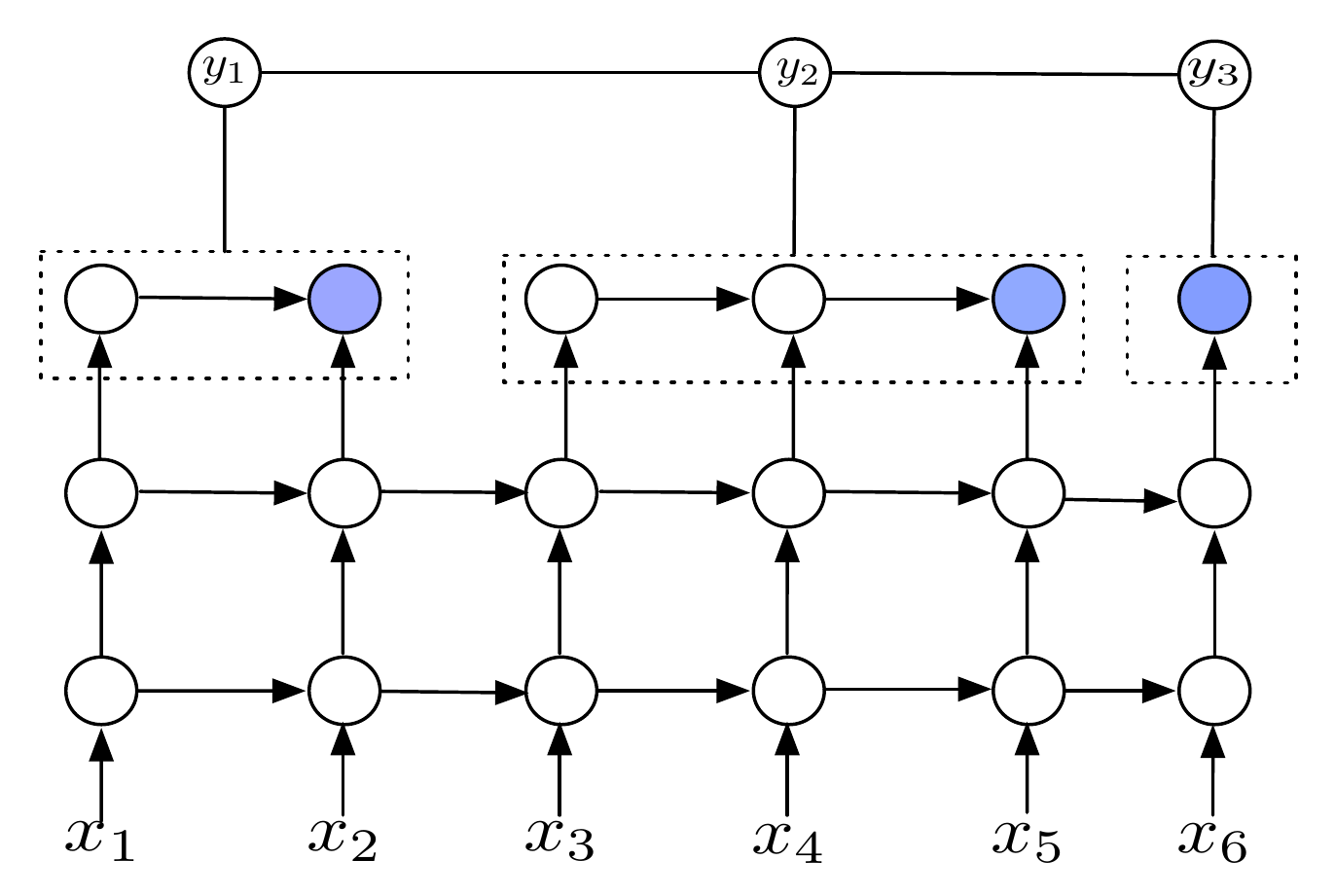}} \vskip -4mm
\caption{Segmental RNN using a first-order CRF. The coloured circles denote the segment embedding vector $\mathbf{h}_{d_j}^j$ in Eq.\eqref{eq:emb}. Using bi-directional RNNs is straightforward. }  
\label{fig:segrnn}
\vskip -5mm
\end{figure}

We use neural networks to define the feature function $\Phi(\cdot)$, which maps the acoustic segment and its corresponding label into a joint feature space. More specifically, $y_j$ is firstly represented as a one-hot vector $\mathbf{v}_j$, and it is then mapped into a continuous space by a linear embedding matrix $\mathbf{M}$ as
\begin{align}
\mathbf{u}_j = \mathbf{Mv}_j
\end{align}
Given the segment label $\mathbf{e}_j$, we use an RNN to map the acoustic segment to a fixed-dimensional vector representation,  i.e.,
\begin{align}
&\mathbf{h}_{1}^j =   r(\mathbf{h}_0, \mathbf{x}_{s_j}) \\
&\mathbf{h}_{2}^j =  r(\mathbf{h}_1^j, \mathbf{x}_{s_j + 1}) \\
& \quad \vdots \nonumber \\ 
\label{eq:emb}
&\mathbf{h}_{d_j}^j   =  r(\mathbf{h}_{d_j -1}^j, \mathbf{x}_{n_{j}}) 
\end{align}
where $\mathbf{h}_0$ denotes the initial hidden state, $d_j = n_{j} - s_{j}$ denotes the duration of the segment and $r(\cdot)$ is a non-linear function. We take the final hidden state $\mathbf{h}_{d_j}^j$ as the segment embedding vector, then $\Phi(\cdot)$ can be represented as 
\begin{align}
\Phi(y_j, \mathbf{e}_j, \mathbf{X}) = g(\mathbf{u}_j, \mathbf{h}_{d_j}^j),
\end{align}
where $g(\cdot)$ corresponds to one layer or multiple layers of linear or non-linear transformation. In fact, it is flexible to include other relevant features as additional inputs to the function $g(\cdot)$, e.g., the duration feature which can be obtained by converting $d_j$ into another embedding vector. In practice, multiple RNN layers can be used transform the acoustic signal $\mathbf{X}$ before extracting the segment embedding vector $\mathbf{h}_{d_j}^j$ as Figure \ref{fig:segrnn}. 

\subsection{Conditional Maximum Likelihood Training}

For speech recognition, the segmentation labels $\mathbf{E}$ are usually unknown,  training the model by maximising the conditional probability as Eq. \eqref{eq:crf} is therefore not practical. The problem can be addressed by defining the loss function as the negative marginal log-likelihood as
\begin{align}
\mathcal{L}(\theta) &= - \log P(\mathbf{y} \mid \mathbf{X}) \nonumber \\
 &= - \log \sum_{\mathbf{E}} P(\mathbf{y, E} \mid \mathbf{X}) \nonumber \\
&= - \log \underbrace{\sum_{\mathbf{E}} \prod_j \exp f \left( y_j, \mathbf{e}_j, \mathbf{X} \right)}_{\equiv Z(\mathbf{X}, \mathbf{y})} + \log Z(\mathbf{X}),
\end{align}
where $\theta$ denotes the set of model parameters, and $Z(\mathbf{X}, \mathbf{y})$ denotes the summation over all the possible segmentations when only $\mathbf{y}$ is observed. To simplify notations, the objective function $\mathcal{L}(\theta)$ is define with only one training utterance. 

However, the number of possible segmentations is exponential with the length of $\mathbf{X}$, which makes the naive computation of both $Z(\mathbf{X}, \mathbf{y})$ and $Z(\mathbf{X})$ impractical. Fortunately, this can be addressed by using the following dynamic programming algorithm as proposed in~\cite{sarawagi2004semi}:
\begin{align}
\alpha_0 &= 1 \\
\label{eq:alpha}
\alpha_{t} & = \sum_{0<k<t} \alpha_{k} \times \sum_{y\in \mathcal{Y}} f(y, \langle k, t \rangle, \mathbf{X}) \\
Z(\mathbf{X}) & = \alpha_T
\end{align}
In Eq. \eqref{eq:alpha}, the first summation is over all the possible segmentation up to timestep $t$, and the second summation is over all the possible labels from the vocabulary. The computation cost of this algorithm is $O(T^2\cdot |\mathcal{Y}|)$, where $|\mathcal{Y}|$ is the size of the vocabulary. The cost can be further reduced by introducing an upper bound of  the segment length, in which case Eq. \eqref{eq:alpha} can be rewritten as
\begin{align}
\label{eq:alpha2}
\alpha_{t} & = \sum_{l<k<t} \alpha_{k} \times \sum_{y\in \mathcal{Y}} f(y, \langle k, t \rangle, \mathbf{X}) \\
\label{eq:seglen}
l & = \left \{ \begin{array}{l}
0 \qquad \quad \text{if} \quad t - L < 0 \\
t - L \qquad \text{otherwise} \\
\end{array}
\right .
\end{align}
where $L$ denotes the maximum value of the segment length. The cost is then reduced to $O(L\cdot T\cdot |\mathcal{Y}|)$, and for long sequences like speech signals where $T \gg L$, the computational savings are substantial. 

The term $Z(\mathbf{X}, \mathbf{y})$ can be computed similarly. In this case, since the label $\mathbf{y}$ is now observed, the summation over all the possible labels $y \in \mathcal{Y}$ in Eq. \eqref{eq:alpha} is not necessary, i.e.,
\begin{align}
\beta_{0,0} &= 1 \\
\label{eq:beta}
\beta_{t,j} & = \sum_{0<k<t} \beta_{k,j-1} \times  f(y_j, \langle k, t \rangle, \mathbf{X}) \\
Z(\mathbf{X}, \mathbf{y}) & = \beta_{T,J}
\end{align}
Again, we can limit the length of the possible segments as Eq. \eqref{eq:alpha2}. Given $Z(\mathbf{X})$ and $Z(\mathbf{X}, \mathbf{y})$, the loss function $\mathcal{L}(\theta)$ can be minimised using the stochastic gradient decent (SGD) algorithm similar to training other neural network models. Other losses, for example, hinge, can be considered in future work.

\begin{figure}[t]
\small
\centerline{\includegraphics[width=0.3\textwidth]{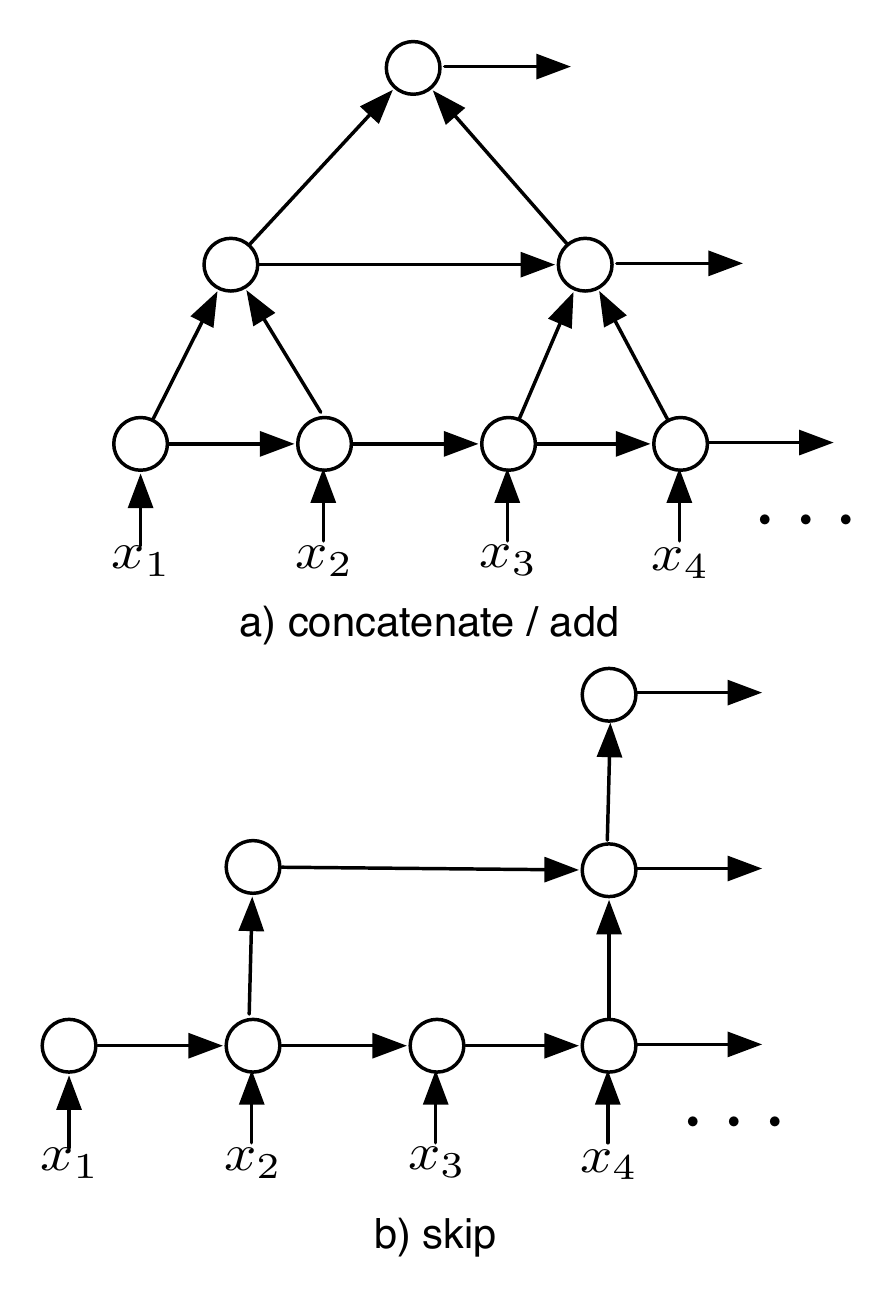}} \vskip -4mm
\caption{ Hierarchical subsampling recurrent network~\cite{graves2012hierarchical} . The size of the subsampling window is two in this example.  }  
\label{fig:hsrnn}
\vskip -5mm
\end{figure}

\subsection{Viterbi Decoding}

During decoding, we need to search the target label sequence $\mathbf{y}$ that yields the highest posterior probability given $\mathbf{X}$ by marginalising out all the possible segmentations:
\begin{align}  
\mathbf{y}^* = \arg\max_{\mathbf{y}} \log \sum_{\mathbf{E}} P(\mathbf{y}, \mathbf{E} \mid \mathbf{X})
\end{align}
This involves minor modification of the recursive algorithm in Eq. \eqref{eq:alpha} that instead of summing over all the possible labels, the Viterbi path up to the timestep $t$ is
\begin{align}
\alpha_{t}^* & = \sum_{0<k<t} \alpha_{k}^* \times \max_{y\in \mathcal{Y}} f(y, \langle k, t \rangle, \mathbf{X}) 
\end{align}
However, marginalising out all the possible segmentations is still expensive. The computational cost can be further reduced by greedy searching the most likely segmentation,  i.e.,
\begin{align}
\alpha_{t}^* & = \max_{0<k<t} \alpha_{k}^* \times \max_{y\in \mathcal{Y}} f(y, \langle k, t \rangle, \mathbf{X}),
\end{align}
which corresponds to the decoding objective as
\begin{align}  
\mathbf{y}^*, \mathbf{E}^*= \arg\max_{\mathbf{y, E}} \log P(\mathbf{y}, \mathbf{E} \mid \mathbf{X})
\end{align}
This joint maximization algorithm may yield high search error, because it only considers one segmentation.  In the future, we shall investigate the beam search algorithm which may yield a lower search error.

\subsection{Further Speedup}
\label{sec:subsample}

It is computationally expensive for RNNs to model long sequences, and the number of possible segmentations is exponential with the length of the input sequence as mentioned before. The computational cost can be significantly reduced by using the hierarchical subsampling RNN~\cite{graves2012hierarchical} to shorten the input sequences, where the subsampling layer takes a window of hidden states from the lower layer as input as shown in Figure \ref{fig:hsrnn}.  In this work, we consider three variants: a) {\it concatenate} -- the hidden states in the subsampling window are concatenated before been fed into the next layer; b) {\it add} -- the hidden states are added into one vector for the next layer; c) {\it skip} -- only the last hidden state in the window is kept and all the others are skipped. The last two schemes are computationally cheaper as they do not introduce extra model parameters. 

\section{Experiments}
\label{sec:exp}

 \begin{table}
 \centering \small
\caption{Speedup by hierarchical subsampling networks. }\vskip 1.5mm

\label{tab:speedup}
\begin{tabular}{l|cc}
\hline

\hline
subsampling   & $L$ &  speedup  \\ \hline
No & 30  & 1 \\
1 layer & 15   & $\sim$3x\\
2 layers & 8 & $\sim$10x \\

\hline

\hline
\end{tabular}
\vskip-4mm
\end{table}

 \begin{table}
 \centering \small
\caption{Results of hierarchical subsampling networks. $d(\mathbf{w})$ and $d(\mathbf{h}_j)$ denote the dimension of $\mathbf{w}$ and $\mathbf{h}_{d_j}^j$ in Eqs. \eqref{eq:phi} and \eqref{eq:emb} respectively. {\tt layers} denotes the number of LSTM layers and {\tt hidden} is the dimension of the LSTM  cells. We reduced the dimension of $\mathbf{h}_{d_j}^j$ from the LSTM output for computational reasons. {\tt conc} is short for the concatenating operation. }\vskip 1.5mm

\label{tab:hsrnn}
\begin{tabular}{l|ccccc}
\hline

\hline
System  & $d(\mathbf{w}$) & $d(\mathbf{h}_{d_j}^j)$ & layers & hidden &  PER(\%) \\ \hline
{\tt skip} & 64 & 64 & 3 & 128 & 21.2 \\
{\tt conc} & 64 & 64 & 3 & 128  & 21.3 \\
%{\tt conc} & 64 & 32 & 3 & 128  & 23.9 \\
{\tt add}  & 64 & 64 & 3 & 128  & 23.2 \\ \hline
{\tt skip} & 64 & 64 & 3 & 250 & 20.1 \\
{\tt conc} & 64 & 64 & 3 & 250 & 20.5 \\ 
{\tt add} & 64 & 64 & 3 & 250 & 21.5  \\

\hline

\hline
\end{tabular}
\vskip-5mm
\end{table}

\subsection{System Setup}

We used the TIMIT dataset to evaluate the segmental RNN acoustic models. This dataset was preferred for the rapid evaluation of different system settings, and for the comparison to other CRF and end-to-end systems. We followed the standard protocol of the TIMIT dataset, and our experiments were based on the Kaldi recipe~\cite{povey2011kaldi}. We used the core test set as our evaluation set, which has 192 utterances. We used 24 dimensional log fiterbanks (FBANKs) with delta and double-delta coefficients, yielding 72 dimensional feature vectors. Our models were trained with 48 phonemes, and their predictions were converted to 39 phonemes before scoring. The dimension of $\mathbf{u}_j$ was fixed to be 32. For all our experiments, we used the long short-term memory (LSTM) networks~\cite{hochreiter1997long} as the implementation of RNNs, and the networks were always bi-directional. We set the initial SGD learning rate to be 0.1, and we exponentially decay the learning rate by a factor of 2 when the validation error stopped decreasing. Our models were trained with dropout regularisation~\cite{srivastava2014dropout}, using an specific implementation for recurrent networks~\cite{zaremba2014recurrent}. The dropout rate was 0.2 unless specified otherwise. Our models were randomly initialised with the same random seed.

\subsection{Results of Hierarchical Subsampling}

We first demonstrate the results of the hierarchical subsampling recurrent network, which is the key to speed up our experiments. We set the size of the subsampling window to be 2, therefore each subsampling layer reduced the time resolution by a factor of 2. We set the maximum segment length $L$ in Eq. \eqref{eq:seglen} to be 300 milliseconds, which corresponded to 30 frames of FBANKs (sampled at the rate of 10 milliseconds).  With two layers of subsampling recurrent networks, the time resolution was reduced by a factor of 4, and the value of $L$ was reduced to be 8, yielding around 10 times speedup as shown in Table \ref{tab:speedup}. 

Table \ref{tab:hsrnn} compares the three implementations of the recurrent subsampling network detailed in section \ref{sec:subsample}. We observed that concatenating all the hidden states in the subsampling window did not yield lower phone error rate (PER) than using the simple {\it skipping} approach, which may be due to the fact that the TIMIT dataset is small and it prefers a smaller model. On the other hand, adding the hidden states in the subsampling window together worked even worse, possibly due to that the sequential information in the subsampling window was flattened. In the following experiments, we sticked to the {\it skipping} method, and using two subsampling layers.

 \begin{table}
 \centering \small
\caption{Results of tuning the hyperparameters. }\vskip 1.5mm
\label{tab:tune}
\begin{tabular}{c|cccccc}
\hline

\hline
Dropout  & $d(\mathbf{w}$) & $d(\mathbf{h}_{d_j}^j)$ & layers & hidden & PER  \\ \hline
 & 64 & 64 & 3 & 128  & 21.2 \\
%SegRNN-dr0.1 & 64 & 32 & 3 & 128 & No & 24.2 \\
& 64 & 32 & 3 & 128 & 21.6 \\
 & 32 & 32 & 3 & 128  & 21.4 \\ \cline{2-6}
  & 64 & 64 & 3 & 250 & 20.1 \\
 0.2 & 64 & 32 & 3 & 250 & 20.4 \\
 & 32 & 32 & 3 & 250 & 20.6 \\ \cline{2-6}
 & 64 & 64 & 6 & 250 & 19.3 \\
 & 64 & 32 & 6 & 250 & 20.2  \\
 & 32 & 32 & 6 & 250 & 20.2 \\ \hline
 & 64 & 64 & 3 & 128 & 21.3 \\
0.1 & 64 & 64 & 3 & 250 & 20.9 \\
 & 64 & 64 & 6 & 250 & 20.4 \\ \hline
$\times$  & 64 & 64 & 6 & 250 & 21.9 \\
\hline

\hline
\end{tabular}
\vskip-4mm
\end{table}

\begin{table}
 \centering \small
\caption{Results of three types of acoustic features. }\vskip 1.5mm
\label{tab:feautre}
\begin{tabular}{l|cccccc}
\hline

\hline
Features  & Deltas & $d(\mathbf{x}_t)$ & PER  \\ \hline
 24-dim FBANK & $\surd$ & 72 & 19.3 \\
 40-dim FBANK & $\surd$ & 120 & 18.9 \\
 Kaldi & $\times$ & 40 &  17.3 \\ 
 \hline

\hline
\end{tabular}
\vskip-5mm
\end{table}

\subsection{Hyperparameters and Different Features}

We then evaluated the model by tuning the hyperparameters, and the results are given in Table \ref{tab:tune}. We tuned the number of LSTM layers, and the dimension of LSTM cells, as well as the dimensions of $\mathbf{w}$ and the segment vector $\mathbf{h}_{d_j}^j$. In general, larger models with dropout regularisation yielded higher recognition accuracy. Our best result was obtained using 6 layers of 250-dimensional LSTMs. However, without the dropout regularisation, the model can be easily overfit due to the small size of training set. In the future, we shall evaluate this model with a large dataset. 

We then evaluated another two types of features using the same system configuration that achieved the best result in Table \ref{tab:tune}. We increased the number of FBANKs from 24 to 40, which yielded slightly lower PER. We also evaluated the standard Kaldi features --- 39 dimensional MFCCs spliced by a context window of 7, followed by LDA and MLLT transform and with feature-space speaker-dependent MLLR, which were the same features used in the HMM-DNN baseline in Table \ref{tab:compare}. The well-engineered features improved the accuracy of our system by more than 1\% absolute.

 \begin{table}
 \centering \small
\caption{Comparison to Related Works. {\tt LM} denotes the language model, and {\tt SD} denotes speaker-dependent transform. The HMM-DNN baseline was trained with cross-entropy using the Kaldi recipe. Sequence training did not improve it due to the small amount of data. Note that RNN transducer and attention-based RNN are equipped with built-in RNNLMs.}\vskip 1.5mm

\label{tab:compare}
\begin{tabular}{l|ccc}
\hline

\hline
System  & LM & SD & PER  \\ \hline
HMM-DNN & $\surd$ & $\surd$  & 18.5 \\
%HMM-DNN (sMBR) & $\surd$ & $\surd$  &  \\
first-pass SCRF~\cite{zweig2012classification} & $\surd$ & $\times$ & 33.1  \\
Boundary-factored SCRF~\cite{he2012efficient} & $\times$  & $\times$ & 26.5 \\
Deep Segmental NN~\cite{abdel2013deep} & $\surd$ & $\times$ & 21.9 \\
Discriminative segmental cascade~\cite{tang2015discriminative} & $\surd$ & $\times$ & 21.7 \\
$\quad$ + 2nd pass with various features & $\surd$ & $\times$ & 19.9 \\ \hline
CTC~\cite{graves2013speech} & $\times$ & $\times$ & 18.4 \\
RNN transducer~\cite{graves2013speech} & -- & $\times$ & 17.7 \\
Attention-based RNN~\cite{chorowski2015attention} & -- & $\times$ & 17.6 \\
 %+Conv. Features + Smooth Focus~\cite{chorowski2015attention}  & $\times$ & $\times$ & 17.6 \\
Segmental RNN &$\times$ & $\times$ & 18.9 \\
Segmental RNN &$\times$ & $\surd$ & 17.3 \\
 
\hline

\hline
\end{tabular}
\vskip-4mm
\end{table}

\subsection{Comparison to Related Works }

In Table \ref{tab:compare}, we compare our result to other reported results using segmental CRFs as well as recent end-to-end systems. Previous state-of-the-art result using segmental CRFs on the TIMIT dataset is reported in~\cite{tang2015discriminative}, where the first-pass decoding was used to prune the search space, and the second-pass was used to re-score the hypothesis using various features including neural network features. Besides, the ground-truth segmentation was used in~\cite{tang2015discriminative}. We achieved considerably lower PER with first-pass decoding, despite the fact that our CRF was zeroth-order, and we did not use any language model. Furthermore, our results are also comparable to that from the CTC and attention-based RNN end-to-end systems. The accuracy of segmental RNNs may be further improved by using higher-order CRFs or incorporating a language model into the decode step, and using beam search to reduce the search error.

\section{Conclusions}
In this paper, we present the segmental RNN --- a novel acoustic model that combines the segmental CRF with an encoder RNN for end-to-end speech recognition. We discuss the practical training and decoding algorithms of this model for speech recognition, and the subsampling network to reduce the computational cost. Our experiments were performed on the TIMIT dataset, and we achieved strong recognition accuracy using zeroth-order CRF, and without using any language model. In the future, we shall investigate discriminative training criteria, and incorporating a language model into the decoding step. Future works also include implementing a weighted finite sate transducer (WFST) based decoder and scaling this model to large vocabulary datasets.  

\newpage
\small
\bibliographystyle{IEEEtran}
\bibliography{bibtex}

\end{document}